\documentclass[letterpaper, 10 pt, conference]{ieeeconf}
\IEEEoverridecommandlockouts    
\usepackage{glossaries}
\newacronym{mav}{UAV}{Unmanned Aerial Vehicle}
\newacronym{uav}{UAV}{Unmanned Aerial Vehicle}
\newacronym{ovc}{OVC}{Open Vision Computer}
\newacronym{lidar}{LiDAR}{Light Detection and Ranging}
\newacronym{vio}{VIO}{visual-inertial odometry}
\newacronym{satnav}{GNSS}{Global Navigation Satellite Systems}
\newacronym{dgps}{DGPS}{Differential GPS}
\newacronym{rtk}{RTK}{Real-Time Kinematics}
\newacronym{ppk}{PPK}{Post-Processed Kinematic}
\newacronym{gpgpu}{GPGPU}{General-Purpose Graphics Processing Unit}
\newacronym{hri}{HRI}{Human-Robot Interaction}
\newacronym{ugv}{UGV}{Unmanned Ground Vehicle}
\newacronym{uwb}{UWB}{Ultra Wideband}
\newacronym{svm}{SVM}{Support Vector Machine}
\newacronym{fcn}{FCN}{Fully Convolutional Network}
\newacronym{cnn}{CNN}{Convolutional Neural Network}
\newacronym{loam}{LOAM}{LiDAR Odometry and Mapping}
\newacronym{sloam}{SLOAM}{Semantic LiDAR Odometry and Mapping}
\newacronym{slam}{SLAM}{Simultaneous Localization and Mapping}
\newacronym{swap}{SWaP}{Size, Weight, and Power}
\newacronym{iot4ag}{IoT4Ag}{NSF Engineering Research Center for the Internet of Things for Precision Agriculture}
\newacronym{grasp-lab}{GRASP Lab}{the General Robotics, Automation, Sensing and Perception Laboratory}
\newacronym{jps}{JPS}{Jump Point Search}
\newacronym{ukf}{UKF}{Unscented Kalman Filter}
\newacronym{sam}{SAM}{Smoothing and Mapping}
\newacronym{icp}{ICP}{Iterative Closest Point}
\newacronym{imu}{IMU}{Inertial Measurement Unit}
\newacronym{tsdf}{TSDF}{Truncated Signed Distance Field}
\newacronym{esdf}{ESDF}{Euclidean Signed Distance Field}
\newacronym{rrt}{RRT}{A rapidly exploring random tree}
\newacronym{mslam}{MS-SLAM}{Metric-Semantic SLAM}
\newacronym{iou}{IoU}{Intersection over Union}

\let\labelindent\relax

\usepackage[table,usenames,dvipsnames]{xcolor}      
\usepackage{extarrows}                               
\usepackage{enumitem}
\usepackage{wrapfig}

\usepackage{booktabs}
\usepackage{multirow, multicol}
\usepackage{array}
\usepackage{siunitx}
\newcolumntype{P}[1]{>{\centering\arraybackslash}p{#1}}
\newcolumntype{M}[1]{>{\centering\arraybackslash}m{#1}}
\usepackage{booktabs}
\newcolumntype{N}{>{\centering\arraybackslash}m{.5in}}
\newcolumntype{G}{>{\centering\arraybackslash}m{2in}}
\usepackage{amsmath,amssymb,amsfonts,amsthm,dsfont} 
\usepackage{algorithm, algorithmicx, listings}        
\usepackage[noend]{algpseudocode}			              
\makeatletter
\def\BState{\State\hskip-\ALG@thistlm}
\makeatother
\usepackage{graphicx}
\usepackage{tabularx}
\usepackage{subcaption}
\usepackage{textcomp}
\usepackage[font={small}]{caption}   
\usepackage[breaklinks=true, colorlinks, bookmarks=true, citecolor=Black, urlcolor=Violet,linkcolor=Black]{hyperref}
\usepackage{svg}
\usepackage{cite}
\usepackage{amssymb,graphicx}

\newcolumntype{C}[1]{>{\centering\arraybackslash}p{#1}}

\usepackage{cleveref}

\usepackage{everypage}
\usepackage{afterpage}
\usepackage{ifthen}
\usepackage{cleveref}
\usepackage{subcaption}
\crefname{figure}{Fig.}{Figs.}
\crefname{table}{Table.}{Tables.}
\crefname{equation}{Eq.}{Eqs.}

\usepackage[normalem]{ulem}
\usepackage[utf8]{inputenc}
\usepackage[english]{babel}
\usepackage{hyperref}
\hypersetup{
    colorlinks=true,
    linkcolor=blue,
    filecolor=magenta,      
    urlcolor=cyan,
}

\DeclareMathAlphabet\mathbfcal{OMS}{cmsy}{b}{n}

\newtheorem*{assumption*}{Assumption}

\newtheorem*{problem*}{Problem}

\usepackage{lipsum}
\usepackage{svg}
\usepackage{mathtools}

\usepackage{censor}
\def\censorcolor{gray!50} \let\svcensorrule\censorrule \renewcommand\censorrule[1]{ \textcolor{\censorcolor}{\svcensorrule{#1}}}

\begin{document}
\title{\bf 
Learning When to See and When to Feel: Adaptive Vision-Torque Fusion for Contact-Aware Manipulation
}

\author{Jiuzhou Lei$^{1}$, Chang Liu$^{1}$, Yu She$^{2}$, Xiao Liang$^{3}$, and Minghui Zheng$^{1}$
\thanks{This work was supported by the U.S. National Science Foundation under Grant No. 2422826 and 2527316.}
\thanks{$^{1}$ Jiuzhou Lei, Chang Liu, and Minghui Zheng are with the J. Mike Walker '66 Department of Mechanical Engineering, Texas A\&M University, College Station, TX 77843, USA. {\tt\small Emails: \{jiuzl, changliu.chris, mhzheng\}@tamu.edu.}}
\thanks{$^{2}$ Yu She is with the Edwardson School of Industrial Engineering, Purdue University, West Lafayette, Indiana, USA.  {\tt\small Email: yushe@purdue.edu}}
\thanks{$^{3}$ Xiao Liang is with the Zachry Department of Civil and Environmental Engineering, Texas A\&M University, College Station, TX 77843, USA. {\tt\small Email: xliang@tamu.edu.}}
\thanks{Project code is available
\href{https://github.com/RollingOat/vision-torque-fusion-for-contact-aware-manipulation.git}{here}.}
}

\maketitle

\begin{abstract}

Vision-based policies have achieved a good performance in robotic manipulation due to the accessibility and richness of visual observations. However, purely visual sensing becomes insufficient in contact-rich and force-sensitive tasks where force/torque (F/T) signals provide critical information about contact dynamics, alignment, and interaction quality. Although various strategies have been proposed to integrate vision and F/T signals, including auxiliary prediction objectives, mixture-of-experts architectures, and contact-aware gating mechanisms, a comparison of these approaches remains lacking. 
In this work, we provide a controlled comparison of different F/T--vision integration strategies within diffusion-based manipulation policies.
In addition, we propose an adaptive integration strategy that ignores F/T signals during non-contact phases while adaptively leveraging both vision and torque information during contact. Experimental results demonstrate that our method outperforms the strongest baseline by 14\% in success rate, highlighting the importance of contact-aware multimodal fusion for robotic manipulation.
\end{abstract}

\section{Introduction}
\label{sec:introduction}

\begin{figure}[!t]
  \centering
  \includegraphics[width=\columnwidth, trim=0cm 0cm 2.5cm 0cm, clip]{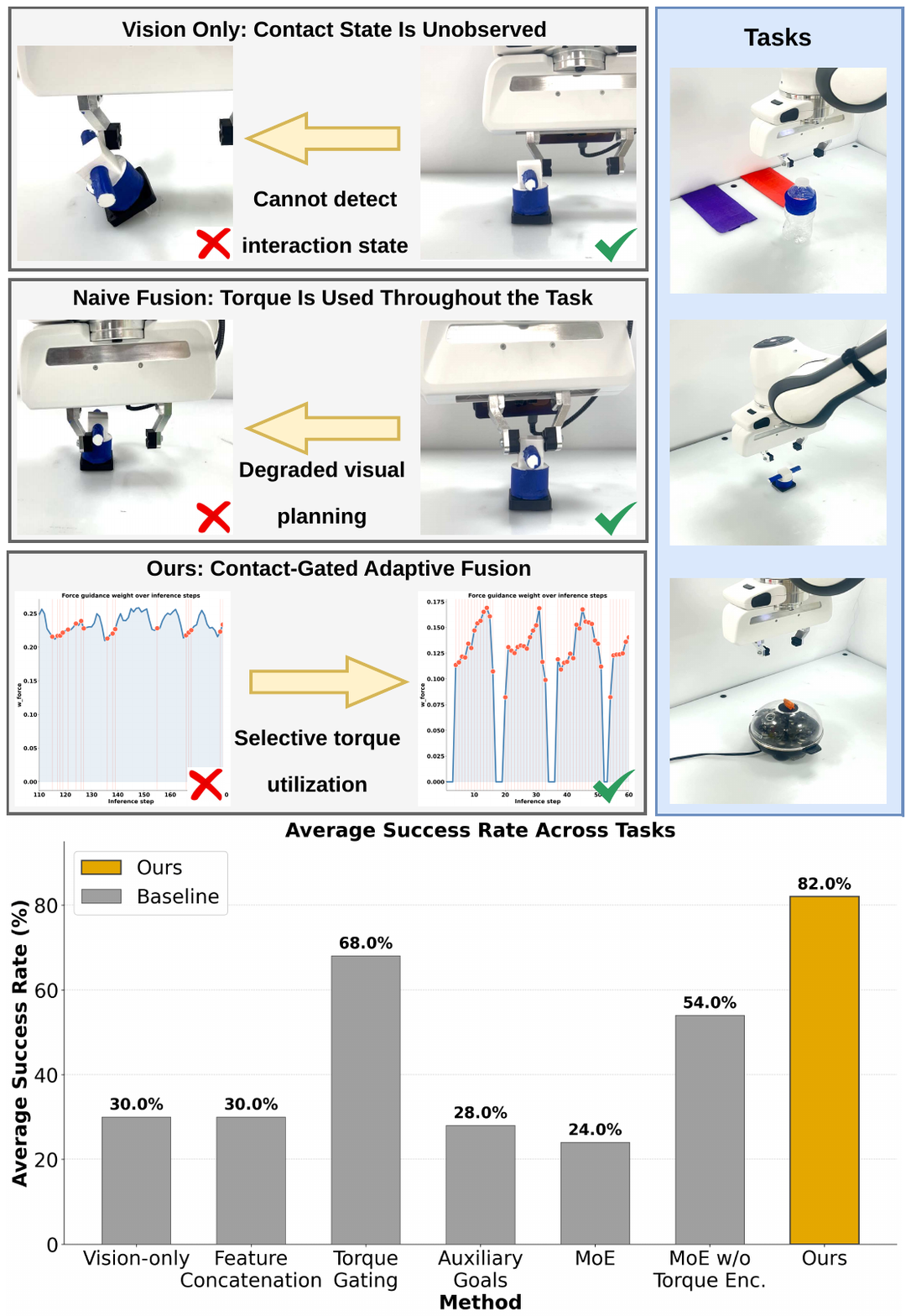}  \caption{\textbf{Contact-Aware Manipulation Challenges.} The top row shows that relying solely on visual input leaves the policy contact-unaware, causing failures in relevant tasks. 
  The middle row shows that naively fusing F/T signals with visual features can degrade policy accuracy during free-space motion. 
  The bottom-left panel shows that the predicted torque guidance weight varies interpretably over inference steps, rising during contact phases and falling during free-space motion. We evaluate across three contact-rich tasks (right), and our method achieves 82\% average success rate, outperforming all baselines by a substantial margin (bottom).}
  \label{fig: Hybrid Vision-Torque Fusion overview}
\end{figure}

Visual cues are sufficient for many daily tasks, and because visual data is relatively easy to acquire, robot learning for manipulation has relied predominantly on image-based sensing \cite{chi2025diffusion, team2024octo, intelligence2025pi_, bjorck2025gr00t}. However, purely visual observations are insufficient in contact-rich and force-aware tasks \cite{tsuji2025survey, suomalainen2022survey}. 
In these scenarios, external force signals at the end effector or joint torque measurements provide critical information about contact dynamics that vision alone cannot capture. For example, in tasks such as connector plug-in/pull-out \cite{zhang2025ta} or cable manipulation \cite{she2021cable}, force/torque (F/T) feedback is often more informative. In those tasks, the invisible or minimal visual changes before and after insertion make it difficult for a vision-only system to determine alignment quality, insertion depth, or contact state. In contrast, F/T signals clearly reflect contact events, misalignment, friction, and successful seating. 
Moreover, F/T sensing and other sensing modalities help bridge gaps when visual information is compromised by self-occlusion. 
During assembly or disassembly tasks, the robotic gripper could frequently block the camera’s view of the workpiece \cite{nottensteiner2021towards, liu2025vision_case_study}. In such cases, F/T feedback provides reliable contact information even when visual observations are partially or fully obstructed.  
Beyond perception, many real-world manipulation tasks require hybrid force and position control. Tasks such as wiping \cite{tsuji2024adaptive} or polishing a surface \cite{solanes2019robust} demand precise spatial trajectory tracking (position control) while simultaneously maintaining a consistent and regulated contact force. Proper force is required to prevent damage to the surface or the tool. 

However, the effective integration of F/T signals into manipulation policy faces some challenges. Directly concatenating F/T features with visual features is often suboptimal, typically resulting in modality collapse as observed in \cite{liu2025factr,chen2025implicitrdp}. Furthermore, integrating F/T data throughout the entire manipulation process risks introducing sensor noise and inertia-induced F/T measurements during free-motion phases, which can degrade the policy's performance.

Several force/torque–vision integration strategies have been proposed. FACTR \cite{liu2025factr} mitigates modality collapse by applying a curriculum of visual corruption to encourage stronger reliance on force inputs. FoAR \cite{he2025foar} introduces a future contact predictor to modulate the contribution of force features, enabling selective haptic utilization during interaction while maintaining stable free-motion behavior.
Routing and expert-based methods further explore dynamic fusion. Chen et al. \cite{chen2025multi} perform policy-level composition by taking a weighted combination of diffusion noise predictions from modality-specific models, while ForceVLA \cite{yu2025forcevla} employs a force-aware mixture-of-experts module that routes tokens to specialized subnetworks conditioned on both task context and interaction feedback.
In parallel, auxiliary prediction strategies strengthen multimodal representations through additional supervision. TA-VLA \cite{zhang2025ta} injects torque into the decoder and jointly predicts torque signals, while ImplicitRDP \cite{chen2025implicitrdp} aligns visual and force encoders via a shared latent target. The Adaptive Compliance Policy \cite{hou2025adaptive} further predicts stiffness and virtual target poses to maintain appropriate compliance during contact-rich tasks. 

Despite these advancements, most existing studies evaluate their methods only against vision-only policies or simple feature concatenation baselines. It remains unclear which integration strategy is most effective in practice. 
We address this gap through the following contributions:
\begin{itemize}
    \item Benchmarking: We provide a controlled comparison of representative F/T--vision integration strategies within a shared diffusion-policy implementation. The compared methods use the same policy backbone, observation space, and training setup, and differ mainly in how torque and visual information are fused.
    \item Algorithm Improvement: We propose an adaptive integration strategy that suppresses force/torque signals during free motion and dynamically incorporates them during contact. This design reduces noise interference while preserving stable visual planning. The resulting policy achieves higher success rates compared to all the baseline fusion strategies.
\end{itemize}

\section{Related Work}
\label{sec: related work}

\begin{figure*}[h]
    \centering
    \includegraphics[width=0.85\textwidth]{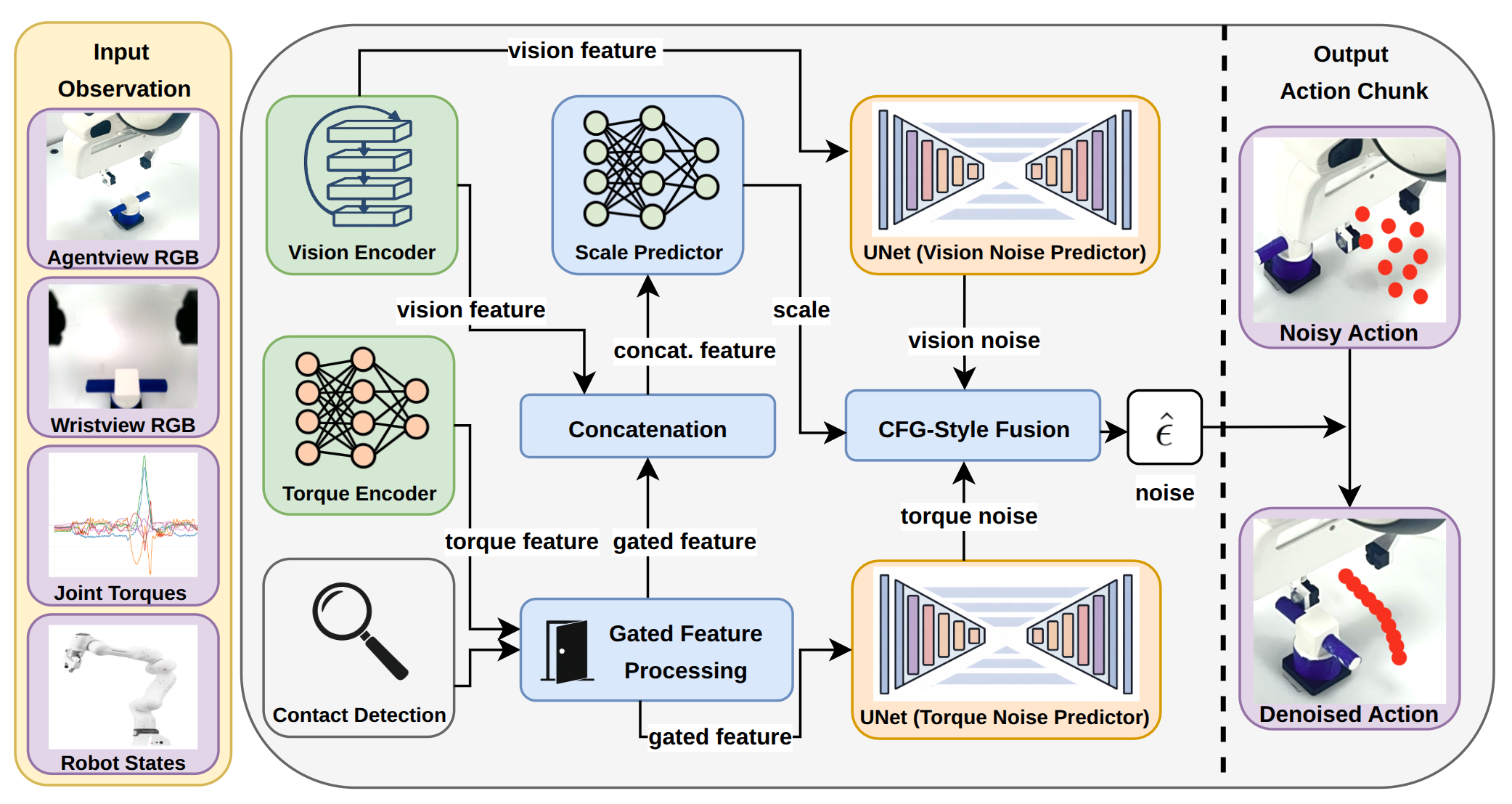}
    \caption{\textbf{Overview of the Proposed Method.} RGB images are encoded using ResNet, with each camera view processed by a separate encoder. Torque signals are encoded using an MLP. The resulting torque features are then passed through a contact-gated module, which modulates them based on contact status to produce contact-gated torque features.
    A scale predictor takes the concatenated vision and torque features as input and outputs a scalar weight that determines the relative influence of torque information. The final denoising noise $\hat{\epsilon}$ is obtained by blending these two predictions according to the predicted scale.}
    \label{fig: methods}
\end{figure*}

Prior work has shown that force and visual feedback can improve contact-rich manipulation compared with vision-only policies \cite{chen2025dexforce, lee2019making, fazeli2019see, liu2025forcemimic, xue2025reactive, kang2025robotic, qi2023general}.
For example, Lee et al. \cite{lee2019making} proposed a self-supervised framework to learn multimodal representations from images, F/T measurements, and proprioceptive states. Their method fuses modalities by jointly predicting optical flow, contact status, and temporal alignment, demonstrating that multimodal inputs significantly improve reinforcement learning sample efficiency in tasks such as peg insertion.
Although not relying on wrist- or joint-mounted F/T sensors, Fazeli et al. \cite{fazeli2019see} presented a hierarchical learning framework integrating vision and tactile sensing for complex manipulation, demonstrated on robotic Jenga. The robot builds a probabilistic generative model that fuses visual and force observations to infer latent physical states and adapt its actions accordingly. By combining multisensory feedback in a structured model, their system achieves more reliable manipulation than vision-only approaches.

Recent studies have shown that naïvely concatenating image features with force/torque (F/T) features can lead to modality collapse and degraded performance due to force sensor noise. Modality collapse refers to the phenomenon in which policies over-rely on high-fidelity visual inputs while ignoring sparse but critical force signals. To mitigate this imbalance, Liu et al. introduced FACTR \cite{liu2025factr}, which applies a curriculum of visual corruption (e.g., Gaussian blur) during training to encourage greater reliance on force inputs. This strategy yields up to a 40\% improvement in success rates on unseen objects in tasks such as dough rolling and box lifting.

Other works explicitly model modality-balancing mechanisms. FoAR \cite{he2025foar} incorporates a future contact predictor to dynamically regulate force utilization, enabling precise control during contact phases while maintaining strong performance in free motion. Chen et al. \cite{chen2025multi} propose a router network to dynamically weight modality-specific experts rather than relying on simple feature-level concatenation, performing policy-level composition by blending noise or score predictions from modality-conditioned diffusion models. Similarly, ForceVLA \cite{yu2025forcevla} introduces a force-aware Mixture-of-Experts (MoE) fusion module that dynamically combines pretrained visual–language representations with real-time 6-axis force-torque feedback during action generation. Its gating network routes tokens to specialized expert subnetworks based on both high-level task instructions and low-level interaction feedback.

Meanwhile, several approaches strengthen force integration through auxiliary objectives. TA-VLA \cite{zhang2025ta} investigates torque integration strategies and finds that injecting torque into the decoder—along with torque prediction as an auxiliary task—leads to more physically grounded representations of interaction dynamics. For tasks requiring explicit force control, controller parameters such as stiffness are predicted jointly with actions. The Adaptive Compliance Policy \cite{hou2025adaptive} further learns to predict a reference pose, a virtual target pose, and a scalar stiffness magnitude to maintain appropriate compliance profiles, demonstrating improved performance in tasks such as pivoting an object against a fixture and wiping a vase. ImplicitRDP \cite{chen2025implicitrdp} introduces a virtual-target representation regularization strategy, where both visual and force encoders are trained to align with a shared modality-agnostic latent target. This auxiliary constraint encourages adaptive modality weighting depending on task conditions and mitigates modality collapse.

Overall, prior strategies for fusing F/T measurements with vision can be broadly categorized into three classes:
(1) \textit{Auxiliary prediction methods}, which learn additional objectives to regularize multimodal representations \cite{chen2025implicitrdp, zhang2025ta, lee2019making};
(2) \textit{Mixture-of-experts (MoE) or routing-based fusion}, which dynamically combine modality-specific experts \cite{chen2025multi, yu2025forcevla};
(3) \textit{Gating mechanisms}, which regulate force usage based on inferred interaction status \cite{he2025foar}.

However, these methods are typically evaluated only against simple baselines such as vision-only policies or naïve feature concatenation. It remains unclear which fusion strategy is most effective in practice. 
To address this gap, this work provides a controlled comparison of representative strategies within a diffusion-policy backbone. 
In addition, we propose an adaptive approach that integrates both contact-gating and classifier-free guidance (CFG) style fusion to enhance policy effectiveness in force/torque-aware tasks.
\section{Proposed Approach}
\label{sec:proposed-approach}

In this section, we explain the core idea of the proposed contact-gated and CFG-style vision–torque fusion. We utilize joint torque measurements as the primary source of force-torque (F/T) information. These torques offer a hardware-efficient and accessible alternative to external sensors, as they are readily available in some robot platforms like Franka Research 3. 

Built upon the diffusion policy\cite{chi2025diffusion} framework, the architecture utilizes a ResNet-18 \cite{he2016deep} as the vision encoder and a Multi-Layer Perceptron (MLP) as the torque encoder, with a U-Net \cite{ronneberger2015u} serving as the noise prediction model. 
The ResNet18 will take in $224 \times 224 \times 3$ RGB images from both the agentview camera and the eye-in-hand view camera. The MLP takes a $7\times10$ vector, which is the joint torque measurements from the past 10 observation steps. 
We chose a history of 10 torque measurements based on the task horizon. This 10-step window is sufficient for the model to perceive both the magnitude and the rate of change in external joint torques.

These encoded features are concatenated to form the conditioning context for the iterative denoising process of the diffusion policy\cite{chi2025diffusion}. Specifically, the noise prediction model takes the current noisy action trajectory and the fused conditioning features as input to estimate the added noise at each diffusion timestep. During inference, the policy generates a sequence of future actions by iteratively refining a trajectory initialized from Gaussian noise over a fixed number of steps. At each denoising step, the CFG-style fusion dynamically weights the contribution of the torque-specialized expert, allowing the model to adapt its action predictions based on the perceived contact state.

In the process, two key components play a vital role in effectively fusing torque and vision information. First, a contact-gating strategy filters out the inertia-induced interference on the torque data during free-space motion. Then a CFG-style vision–torque fusion mechanism allows the policy to learn how to adaptively combine visual and torque modalities based on the current interaction state. The data flow of this approach is visualized in \cref{fig: methods}.

\textbf{Contact Gating.} Following the methodology described in \cite{he2025foar}, we detail the gating mechanism here to ensure the self-containment and readability. The mechanism extends the standard diffusion policy by introducing a contact-aware gate that selectively includes torque information. 

If any of the joints experience external torques larger than a predefined magnitude, it's considered to have contact.
The final force representation is computed as:
\begin{equation}
f_{torque-gated}
=
\phi \cdot f_{torque}
+
(1 - \phi) \cdot f^*
\end{equation}
where $f_{torque}$ is the output of the torque encoder, and ${f}^*$ is a learnable parameter to represent encoded torque features during free-space movement.  When contact is detected ($\phi = 1$), the encoded force features are used directly. When no contact is detected ($\phi = 0$), the force channel is replaced by the learnable parameter ${f}^*$. The gated torque feature is then concatenated with image features for the noise prediction model to condition on.

\textbf{CFG-Style Adaptive Vision-Torque Fusion.}
Rather than employing a single monolithic denoiser, we train two modality-specialized U-Nets. The first, a vision-specialized U-Net, is conditioned on visual and proprioceptive features; the second, a torque-specialized U-Net, is conditioned on gated torque features and proprioceptive features.
Inspired by the classifier-free-guidance diffusion \cite{ho2022classifier}, we inject the noise predicted from the torque-specialized UNet into the noise predicted by the vision-specialized UNet.
The final noise term used for denoising is combined using the classifier-free-guidance-style formulation:
\begin{equation}
\hat{\epsilon}_{\text{final}}
=
\hat{\epsilon}_{\text{vision}}
+
w_{\text{torque}}
\left(
\hat{\epsilon}_{\text{torque}}
-
\hat{\epsilon}_{\text{vision}}
\right).
\end{equation}

When $w_{\text{torque}} = 0$, the output reduces to the pure vision prediction. $w_{\text{scale}}$, the guidance scale, is predicted by a scale predictor implemented as a three-layer MLP. The scale predictor will take the gated torque features and image features as input.  
$w_{\text{torque}}$ itself is modulated by the contact status:
\begin{equation}
\label{eq: scale predictor}
w_{\text{torque}}
=
\phi \cdot \sigma(w_{\text{scale}} ),
\end{equation}
where $\sigma(\cdot)$ denotes the softplus function.

Consequently, the policy exhibits the following behavior:
\begin{itemize}
    \item \textbf{No contact ($\phi = 0$):} $w_{\text{torque}} = 0$, yielding $\hat{\epsilon}_{\text{final}} = \hat{\epsilon}_{\text{vision}}$ and purely vision-driven actions.
    \item \textbf{Contact ($\phi = 1$):} $w_{\text{torque}} \in (0, \infty)$, enabling a learned blend of the vision and torque experts.
\end{itemize}
This design enforces clean separation between free-space and contact regimes while allowing flexible, learned torque amplification during interaction. 

During training, the two modality-specialized UNets and the scale predictor are optimized together using the standard noise prediction objective,
\begin{equation}
\mathcal{L}
=
\mathrm{MSE}(\hat{\epsilon}_{\text{final}}, \epsilon),
\end{equation}
where $\epsilon$ is the sampled ground truth Gaussian noise. 
For reference, \cref{fig: methods} visualizes the modal flow within our proposed architecture.

To better understand the comparison results in \cref{sec:results and analysis}, we also describe the baseline methods in the following subsections.
Since the contact-gating strategy has already been described, we explain only the auxiliary-objective and MoE strategies below.

\subsection{Future-Torque Auxiliary Prediction}
\label{subsec: aux}
 Following the methods in \cite{zhang2025ta}, future torque prediction is used as the auxiliary objective. This strategy augments the standard Diffusion Policy with an auxiliary prediction task on joint torque signals by jointly diffusing robot actions and future torque measurements. The total training loss is computed in noise-prediction space as
\begin{equation}
\mathcal{L}
=
\mathrm{MSE}(\hat{\epsilon}_{\text{action}}, \epsilon_{\text{action}})
+
\alpha \, \mathrm{MSE}(\hat{\epsilon}_{\text{torque}}, \epsilon_{\text{torque}}),
\end{equation}
where $\epsilon_{\text{action}}$ and $\epsilon_{\text{torque}}$ denote the Gaussian noise slices corresponding to action and future torque dimensions. $\hat{\epsilon}_{\text{action}}$ and $\hat{\epsilon}_{\text{torque}}$ are the corresponding UNet predictions, and $\alpha$ is the weighting coefficient set during training. 
At inference time, only the denoised action component is executed; the torque prediction serves only as a training-time regularizer.

\subsection{Modality-Specific Weighted-Average MoE}
\label{subsec: composition}
Following the methodology in \cite{chen2025multi}, we implement a similar Mixture-of-Experts (MoE) baseline for benchmarking as we both do an adaptive fusion on the noise prediction level; however, we simplify the architecture by employing a single U-Net, referred to as a "sub-policy" in \cite{chen2025multi}, for each sensory modality.
Two independent UNets are trained in parallel and their noise predictions are blended at each denoising step via a learned routing network. Instead of conditioning a single UNet on a combined observation, the observation modalities are split into two groups: an image group (RGB images and low-dimensional states such as joint positions) and a torque group. A small MLP is used as the routing network to take the concatenated features from both the image and torque encoders and output two logits that are passed through a softmax to obtain blending weights $(w_{\text{vision}}, w_{\text{tor}})$ that satisfy $w_{\text{vision}} + w_{\text{tor}} = 1$.

During training and inference, the same noisy action trajectory is fed to both UNets. Their predicted noise terms are blended element-wise,
\begin{equation}
\hat{\epsilon}
=
w_{\text{vision}} \cdot \hat{\epsilon}_{\text{vision}}
+
w_{\text{tor}} \cdot \hat{\epsilon}_{\text{torque}},
\label{eq: moe weight}
\end{equation}

\section{Results and Analysis}
\label{sec:results and analysis}

\begin{figure*}[t]
    \centering
    \includegraphics[width=\textwidth, trim=3.5cm 0cm 3.5cm 0cm, clip]{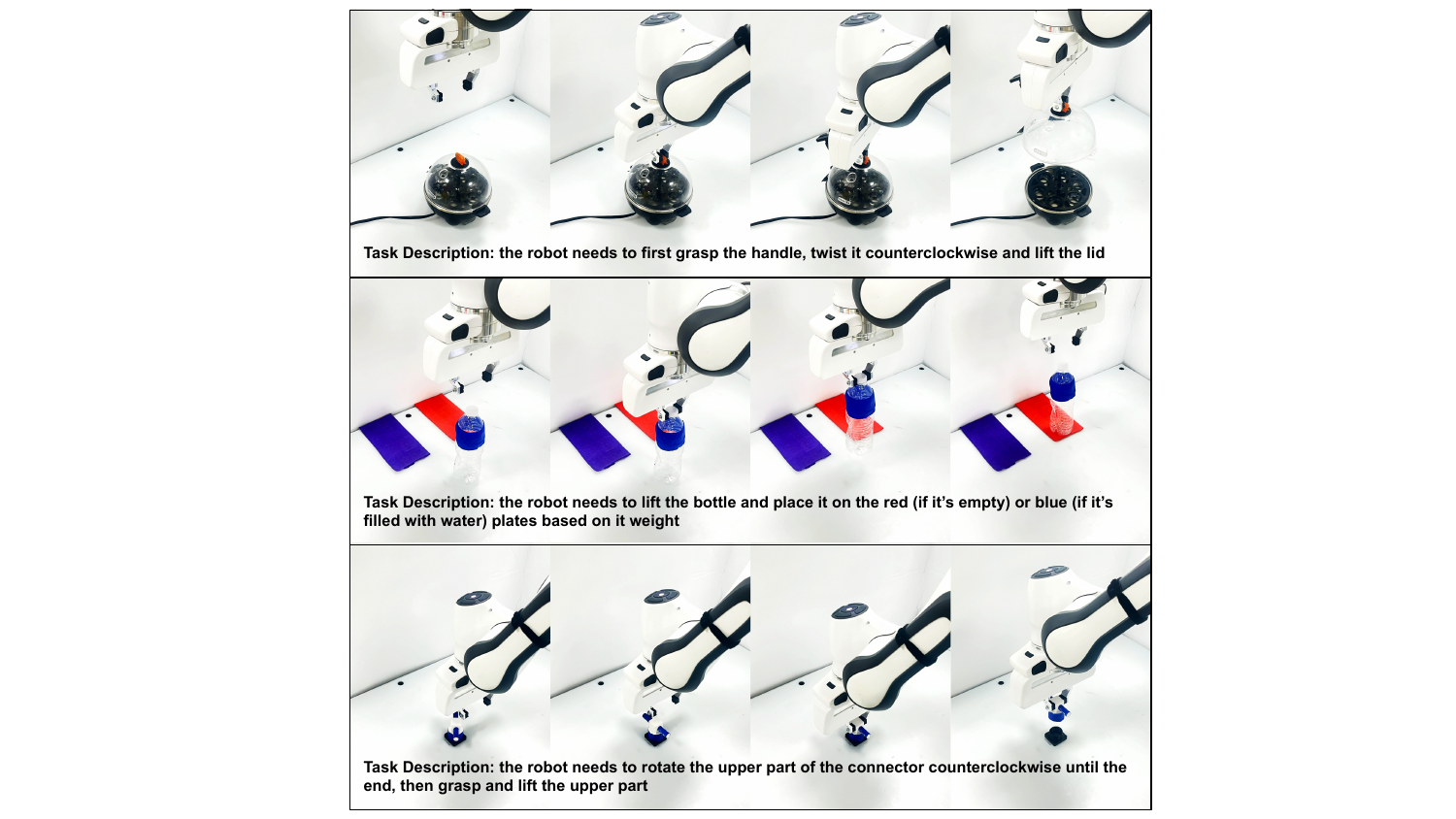}
    \caption{\textbf{Visualization of the Three Tasks.} First row: egg boiler lid opening, second row: weight-based bottle placement, and third row: twisty connector pull out.}
    \label{fig:tasks}
\end{figure*}

\begin{figure}[t]
  \centering
  \includegraphics[width=\columnwidth, trim=3cm 2cm 3cm 0cm, clip]{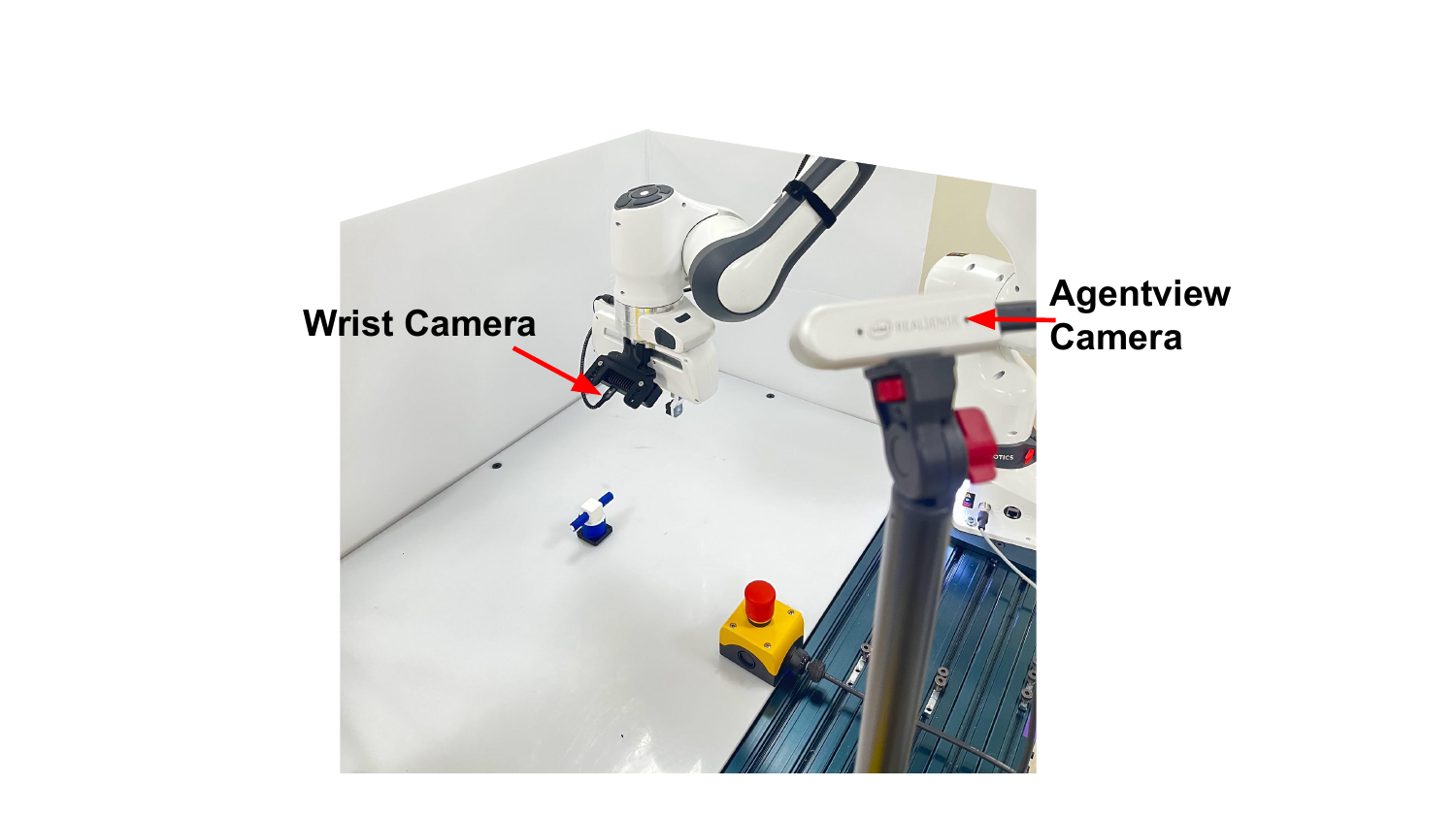}
  \caption{\textbf{Experiment Setup} }
  \label{fig: exp}
\end{figure}

For the experiment section, we will first describe the training, implementation, and used robot platform. Then, the tasks for evaluation will be explained. In addition, the performance of each strategy will be evaluated using success rate per task. Last, we will analyze and discuss the performance of those baselines and our method. 

\subsection{Training and Implementation}
All the algorithms are implemented on top of Robomimic\cite{robomimic2021} in which the original diffusion policy \cite{chi2025diffusion} is implemented and it also provides code interface for the diffusion policy to take multimodal inputs. 
The policy is trained with Robomimic's default setting: optimized with AdamW at an initial learning rate of $1\times10^{-4}$ and a small $L2$ weight decay $(1\times10^{-6})$ for regularization. The diffusion process \cite{ho2020denoising} uses 100 timesteps for both training (forward noising) and inference (denoising). The policy is trained on multimodal data sampled at 10 Hz, incorporating RGB images from both agent-view and wrist-mounted cameras ($224 \times 224 \times 3$), a 70-dimensional vector of joint external torque measurements from multiple observation steps, and a 7-dimensional joint position vector.

We collect data from and deploy the trained policy on a Franka Research 3 robot, which provides readily available joint external torque measurements. An agentview RealSense camera is placed on the left side of the robot arm. An OAK-D camera is used as the wrist camera. Refer to \cref{fig: exp} for the experiment setup.

\subsection{Task Design}

To evaluate the effectiveness of different fusion strategies, we design three real-world manipulation tasks that require varying degrees of contact sensitivity and F/T reasoning:

\begin{itemize}
    \item \textbf{Egg Boiler Lid Opening.} The robot twists and lifts the lid of an egg boiler, requiring detection of resistance changes to avoid early lifting before the twisting phase is completed.
    \item \textbf{Weight-Based Bottle Placement.} The robot must place a visually identical water bottle onto one of two plates depending on its weight, requiring force-based discrimination. The empty bottle should be placed on the red plate, while the full bottle should be placed on the blue plate. 
    \item \textbf{Twisty Connector Pull Out.} The robot must pull out a twist-lock connector that requires axial pulling force after rotating the upper part to its limit. The task evaluates whether the policy can sense the end of rotation through resistance and switch to the pull-out stage accordingly.
    
\end{itemize}

Refer to \cref{fig:tasks} for a visualization of the designed tasks. We collected 110 demonstrations for the weight-based bottle placement task, 250 demonstrations for the twisty connector pull out task, and 150 demonstrations for the egg boiler lid opening task. A task is considered failed if it results in robot motion abortion due to collision or running out of time.

\subsection{Quantitative Results and Analysis}
\label{subsec: result and analysis}

\begin{table}[t]
\centering
\caption{Controlled comparison of torque--vision fusion modules
under a shared Robomimic Diffusion Policy backbone. The methods
differ only in the fusion module. The best performance in terms
of task success rate in each column is highlighted in bold.}
\label{tab:success_rates}
\begin{tabular}{lcccc}
\toprule
\textbf{Method} & \textbf{Bottle} & \textbf{Connector} & \textbf{Lid} & \textbf{Average} \\
\midrule
Vision-only              & 8/20  & 0/10 & 7/20  & 30.0\% \\
Feature Concatenation    & 3/20  & 0/10 & 12/20 & 30.0\% \\
Torque Gating            & 14/20 & 5/10 & 15/20 & 68.0\% \\
Auxiliary Goals          & 6/20  & 1/10 & 7/20  & 28.0\% \\
MoE                      & 5/20  & 1/10 & 7/20  & 24.0\% \\
MoE w/o torque encoding  & 15/20 & 1/10 & 11/20 & 54.0\% \\
Ours                    & \textbf{16/20} & \textbf{7/10} & \textbf{18/20} & \textbf{82.0\%} \\
\bottomrule
\end{tabular}
\end{table}

In \cref{tab:success_rates}, the task success rate for each task over 10 or 20 trials is reported. 
Vision Only uses visual and proprioceptive observations without torque input. Feature Concatenation directly concatenates encoded visual and torque features. Torque gating suppresses torque features during non-contact motion. Auxiliary goals jointly denoises robot actions and future torque measurements during training. MoE combines the predictions of vision and torque-specialized denoisers using routing weights. Finally, Contact-Gated CFG Fusion denotes our proposed method.
Among all the methods, the proposed approach has the highest average success rate. The success rate is improved by 14\% in comparison to the best baseline method. The proposed approach also achieves the highest success rate for each single task. 
Among all the baselines, the simple yet effective strategy is to apply a contact gating mechanism. 

It is noteworthy that even in the absence of torque information, the vision-only baseline has a success rate of 8/20 in the water bottle placement task. This task consists of placing an empty bottle and a full bottle. 
We observe that the vision-only policy exhibits a bias toward placing the bottle on the red plate, despite the training dataset containing an equal distribution of demonstrations for both conditions. 
For the completeness of the results, \cref{tab:water_bottle_task_result} reports the success rates of different strategies in the water bottle placement task under both full and empty bottle conditions. 

\begin{table}[h]
\centering
\caption{Success rates of placing empty and full water bottles.}
\label{tab:water_bottle_task_result}
\begin{tabular}{lccc}
\toprule
\textbf{Method} & \textbf{Empty} & \textbf{Full} & \textbf{Total} \\ 
\midrule
Vision-only             & 8/10 & 0/10 & 8/20 \\
Feature Concatenation   & 1/10 & 2/10 & 3/20 \\
Torque Gating           & 7/10 & 7/10 & 14/20 \\
Auxiliary Goals         & 4/10 & 2/10 & 6/20 \\
MoE                     & 5/10 & 0/10 & 5/20 \\
MoE w/o torque encoding & 9/10 & 6/10 & 15/20 \\
Ours                    & 9/10 & 7/10 & 16/20 \\
\bottomrule
\end{tabular}
\end{table}

In the rest of the section, we will analyze the success and failure reasons for each of the baselines and our method.

\begin{figure}[t]
  \centering
  \includegraphics[width=\columnwidth, trim=2cm 0cm 2cm 0cm, clip]{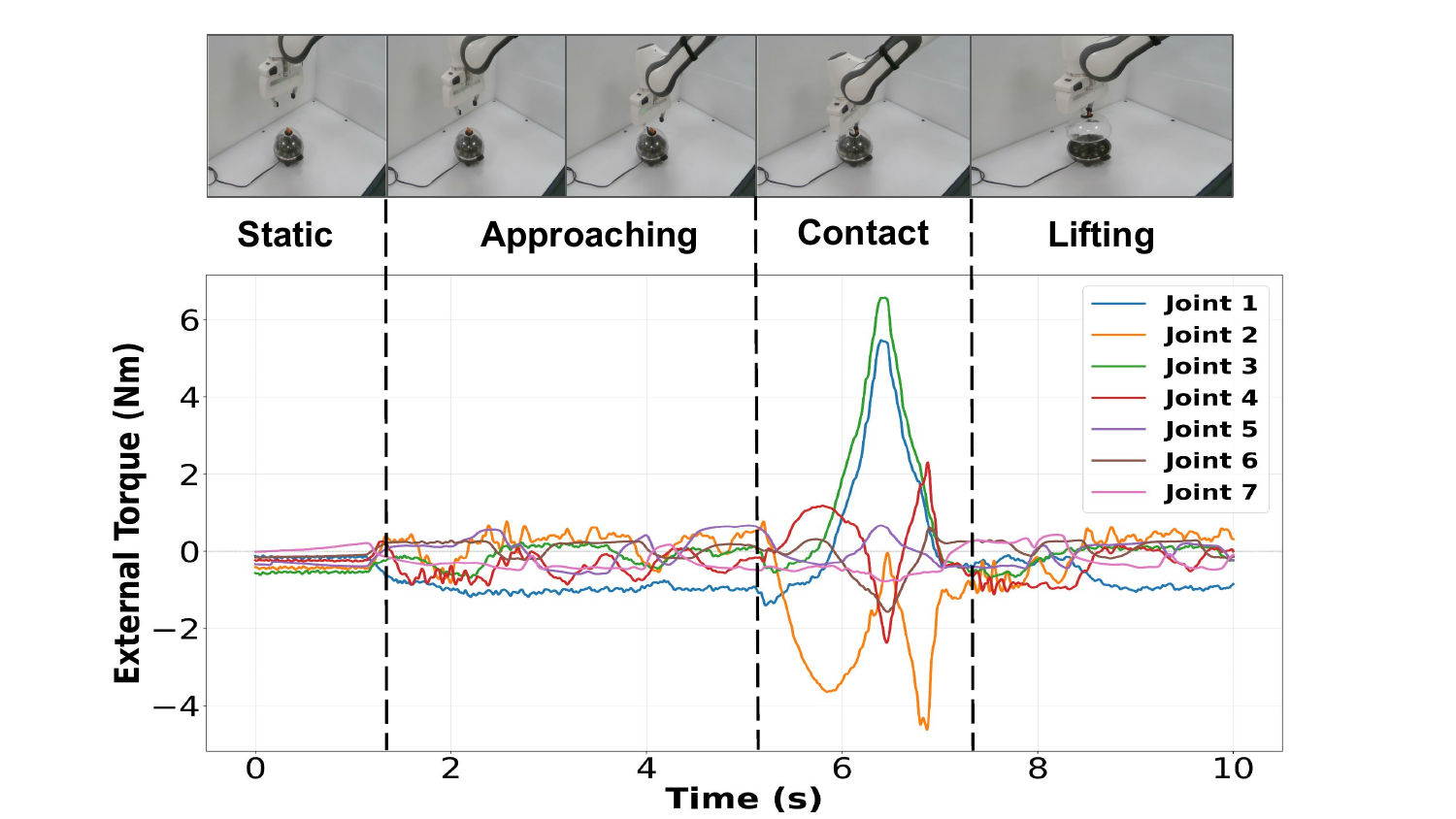}
  \caption{\textbf{External Joint Torque vs. Time During a Task Execution.} During the approaching and lifting stages, the torques measurements are patternless and fluctuate. Different colors represent the torque measurements at different joints.}
  \label{fig: torque vs time}
\end{figure}

\begin{figure}[t]
  \centering
  \includegraphics[width=\columnwidth, trim=0cm 0cm 0cm 0cm, clip]{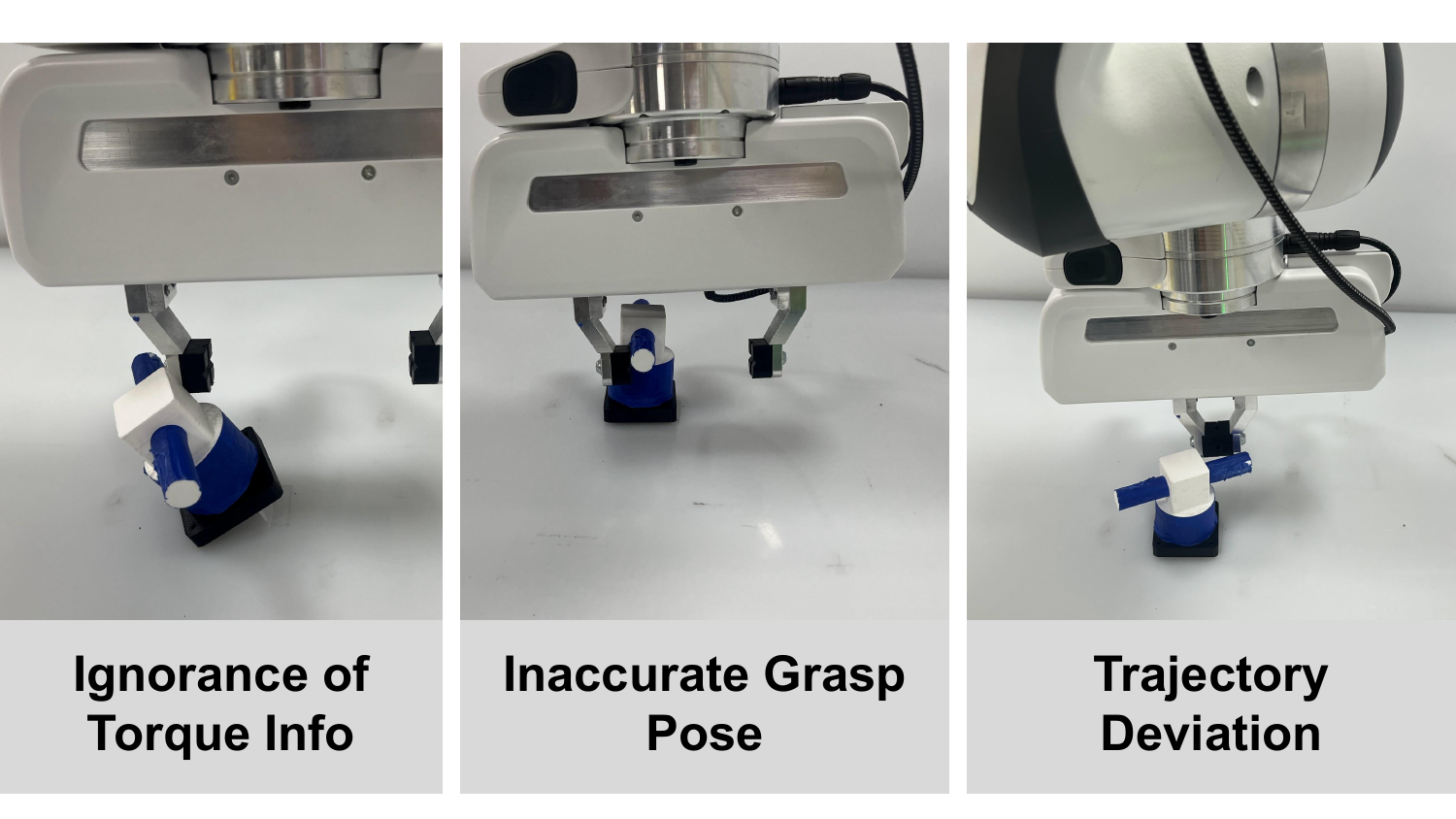}
  \caption{\textbf{Failure Cases Due to Distracting Torque Features}}
  \label{fig: failure cases}
\end{figure}

\textbf{Filtering out torque measurements during free-space motion accounts for the majority of the improvement in success rate.} Unlike dedicated tactile sensors, external joint torques measured during free-space motion are corrupted not only by noise but also by inertial effects. As illustrated in \cref{fig: torque vs time}, which tracks torque fluctuations throughout the entire task execution, these non-contact signals lack a reliable, learnable pattern without an extensively large dataset. Integrating unfiltered torque features can diminish the influence of the image features on which the policy relies for visual planning. This may explain the poor performance of the feature-concatenation method. Though it has a fairly good performance in the task of opening the lid of the egg boiler, it is because the task can tolerate the inaccuracy of grasp pose. It won't lead to motion abortion even when the pose to grasp the handle of the lid deviates from expected poses (the handle is in the middle of the gripper).
In \cref{fig: failure cases}, we show the failure cases of those baselines which do not use contact-gating. As shown in \cref{fig: failure cases}, the failure cases result from inaccurate grasp pose, ignorance of torque information, and deviant trajectory. 

The auxiliary goal strategy exhibits suboptimal performance, which can be partially attributed to the presence of patternless joint torque measurements during free-space movement. While the work in \cite{zhang2025ta} also employs future torque prediction, it utilizes commanded motor torques rather than external torques, which likely provides a cleaner supervisory signal. Furthermore, as noted in \cite{chen2025implicitrdp}, directly predicting future torque sequences may be a difficult auxiliary task given limited amount of data to train a diffusion policy.

\textbf{Learned adaptive fusion of the torque-specialized U-Net enhances manipulation precision.} While the contact-gating baseline achieves competitive success rates across most tasks, it often lacks single-attempt reliability. Observations during experiments indicate that this baseline frequently requires a second trial within the allotted task horizon to achieve the goal. As shown in \cref{tab:single-shot success_rates}, the success rate of opening the lid of the egg boiler in the first attempt along with the average task execution steps are reported. 
Our approach demonstrates higher single-attempt success rates and shorter executed task horizons compared with the contact-gating baseline, suggesting improved interaction precision with the adaptive vision--torque fusion strategy.

\begin{table}[ht]
    \centering
    \caption{Comparison of Single-Shot Success Rates}
    \label{tab:single-shot success_rates}
    \begin{small} 
    \begin{tabular}{l c c} 
        \toprule
        \textbf{Method} & {\textbf{Success Rate (\%)}} & {\textbf{Avg. Task Horizon}} \\ 
        \midrule
        Torque Gating   & 20.0             & 110.7 \\ 
        \textbf{Ours}   & \textbf{60.0}    & \textbf{87.1} \\ 
        \bottomrule
    \multicolumn{3}{l}{\textit{\footnotesize Higher success and lower horizon indicate better performance.}}
    \end{tabular}
    \end{small}
\end{table}

\textbf{Learning adaptive fusion in CFG-style vs. weighted average MoE.} Given that both CFG-style fusion and the Mixture-of-Experts (MoE) baseline integrate vision and torque information at the noise-prediction level in a similar way, we conducted an ablation study to compare their performance on the water bottle placement task. In this experiment, the MoE routing network was modified to receive contact-gated torque features and the torque-specialize UNet is also conditioned on the contact-gated torque features, mirroring the input structure of our proposed method.

\begin{figure}[!htbp]
    \centering
    
    \begin{subfigure}{\columnwidth}
        \centering
        \includegraphics[width=0.85\columnwidth]{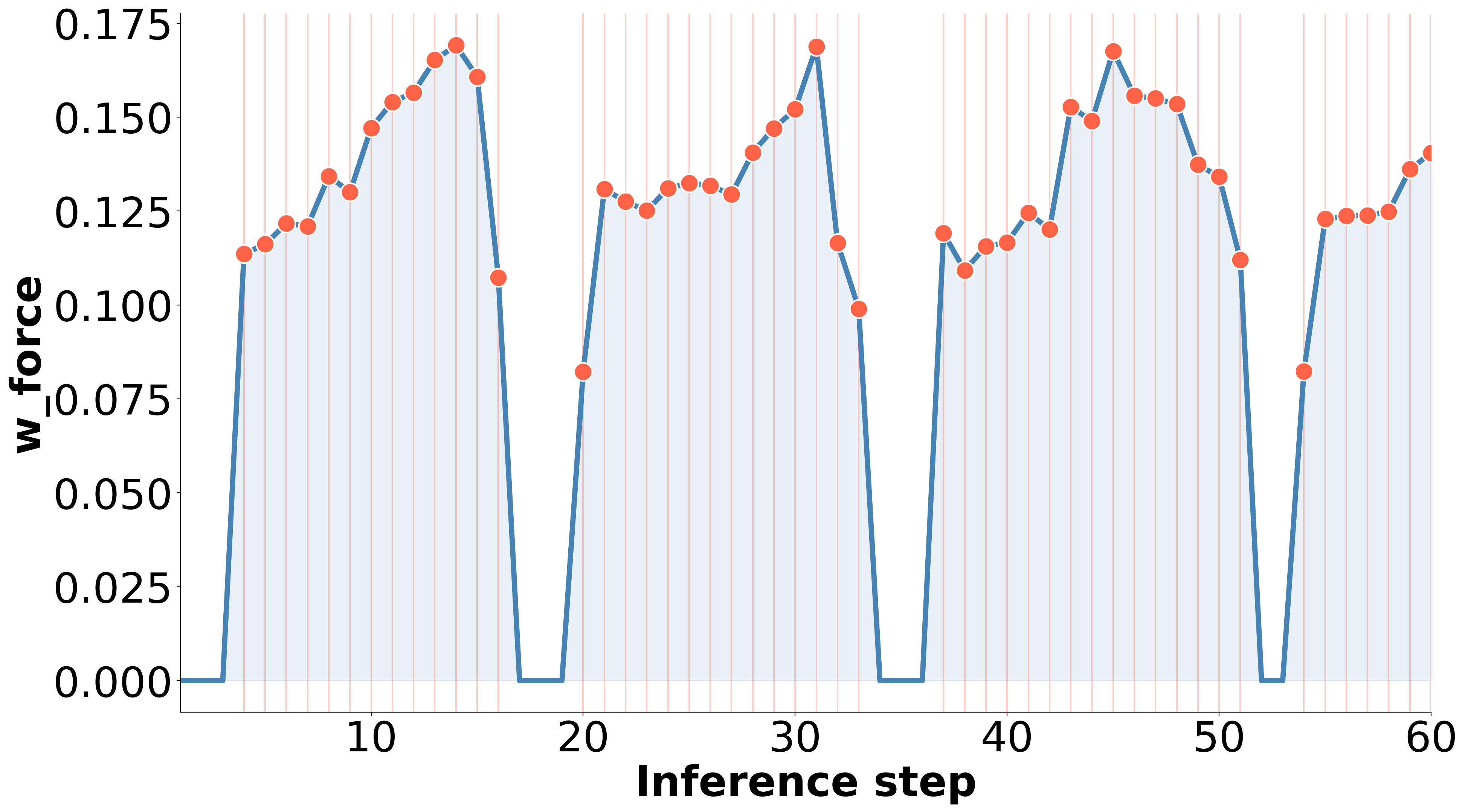}
        \caption{$w_{\text{f/t}}$ values from our approach}
        \label{fig:top_image}
    \end{subfigure}
    
    \vspace{2ex} 
    
    \begin{subfigure}{\columnwidth}
        \centering
        \includegraphics[width=0.85\columnwidth]{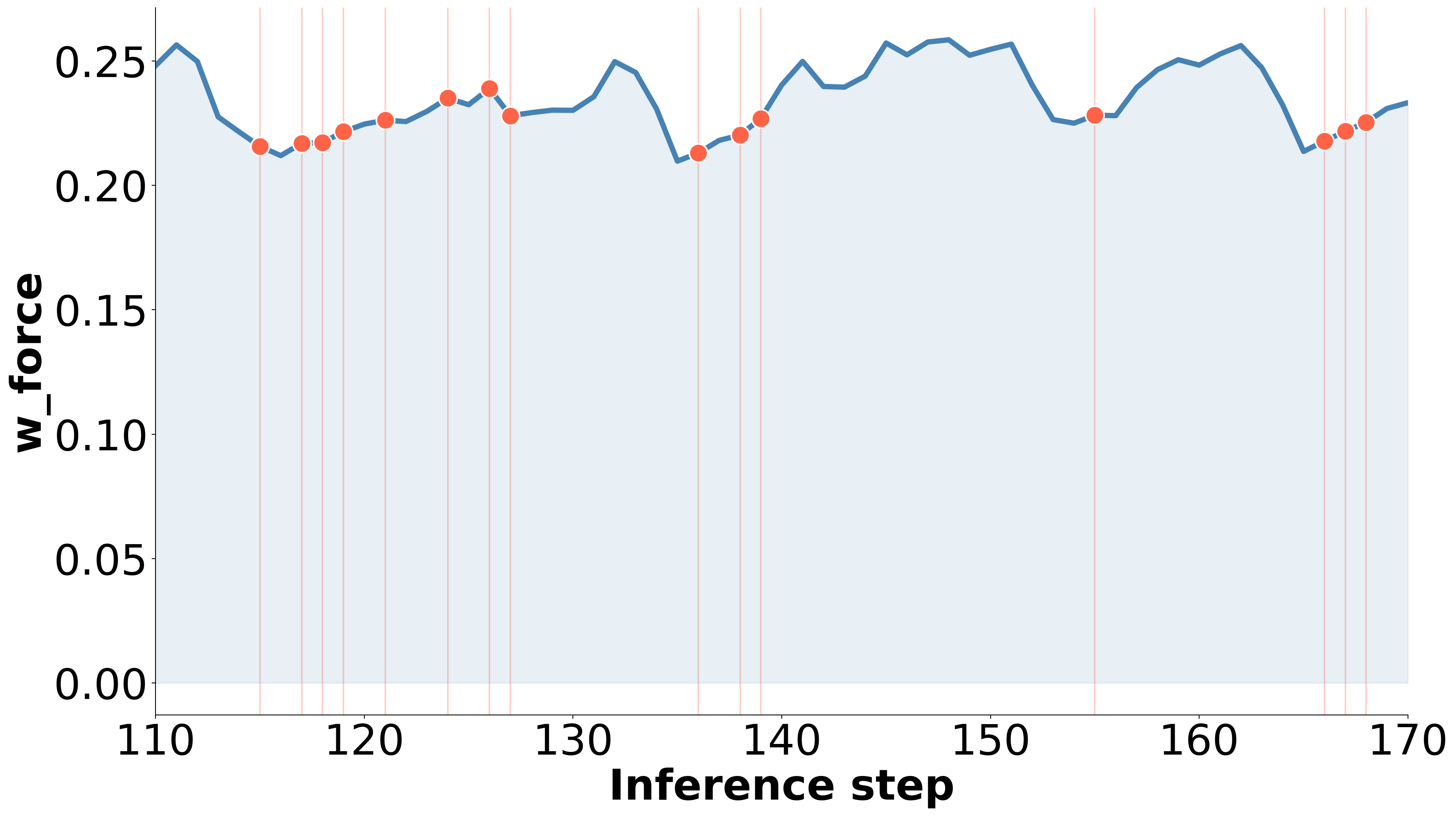}
        \caption{$w_{\text{f/t}}$ values from torque-gated MoE}
        \label{fig:bottom_image}
    \end{subfigure}
    
    \caption{\textbf{Ablation Study.} A comparison between the predicted weights associated with force/torque by our approach (top) and by MoE (bottom) in the ablation study. The orange dots in the figures represent the values of the weights when there is contact.}
    \label{fig:weights}
\end{figure}

Despite the addition of this gating mechanism, the MoE-based method underperforms. An analysis of the routing weights reveals that the model assigns nearly identical importance regardless of the interaction state: during the non-contact phase, the average weights for $w_{\text{vision}}$ and $w_{\text{torque}}$ are 0.7634 and 0.2366, respectively. During contact, these values remain nearly unchanged at 0.7714 and 0.2286. 
\begin{table}[!htbp]
    \centering
    \caption{Ablation Study}
    \label{tab:ablation_study}
    \begin{small}
    \begin{tabular}{l c} 
        \toprule
        \textbf{Method} & {\textbf{Success Rate}} \\ 
        \midrule
        Torque-Gated MoE    & 12/20 \\ 
        \textbf{Ours}   & \textbf{16/20} \\ 
        \bottomrule
    \multicolumn{2}{l}{\textit{\footnotesize Higher success indicates better performance.}}
    \end{tabular}
    \end{small}
\end{table}
In contrast, our proposed approach explicitly regulates the guidance weight $w_{\text{torque}}$ in the scale predictor (\cref{eq: scale predictor}) based on contact status. As illustrated in \cref{fig:weights}, while the MoE weights exhibit an increase during contact, they are indistinguishable from values during free-space motion (represented using blue lines in the figure). 
Conversely, our approach demonstrates high interpretability; the weights vary dynamically and contribute to the policy only when contact is detected (indicated by the orange markers). This ensures that torque information is utilized selectively during physical interaction. 
Without regulating $w_{\text{torque}}$ at the noise prediction level based on contact status, the learned torque features are actually detrimental to the policy because they fail to distinguish between noise and signal, whereas a ``raw'' or ``no-encoder'' approach might work even better as shown in \cref{tab:success_rates}.

These results suggest that the proposed method benefits
from two complementary mechanisms. Contact gating provides
the dominant improvement by preventing unreliable free-space
torque measurements from corrupting visual planning, as shown
by the gap between feature concatenation and the contact-gating
baseline in Table~I. However, gating alone treats the torque
feature as a fixed condition once contact is detected. The
CFG-style adaptive fusion further improves performance by
learning how strongly the torque-specialized denoiser should
modify the vision-specialized prediction during contact. This
explains why the proposed method improves both the overall
success rate and the single-attempt reliability compared with
contact gating alone.

\section{Conclusion and Limitations}

We evaluated representative torque--vision integration strategies within a shared diffusion-policy-based manipulation framework and identified several important design trends. 
The results show that suppressing torque during free-space motion is critical when using external joint torque measurements, and that adaptive torque-conditioned denoising can further improve contact-rich manipulation performance. Across three real-world tasks, the proposed method achieves a 14\% improvement over the strongest baseline. 

Although our evaluation identifies key performance trends, it covers only a representative subset of existing methods. For example, we evaluate an auxiliary-objective strategy based on future torque prediction; however, how to define auxiliary objectives that are both easier to learn and beneficial to policy performance remains underexplored. In addition, the proposed approach currently senses and reasons about F/T data passively. It has been validated only on tasks that do not require the robot to actively apply forces to interact with the environment. Active force control is important for tasks such as surface finishing and precision assembly and disassembly.

\bibliographystyle{IEEEtran}
\bibliography{ref}
\end{document}